\documentclass{article} 
\usepackage{iclr2016_conference,times}
\usepackage{hyperref}
\usepackage{url}
\usepackage{graphicx}
\usepackage{subcaption}
\usepackage{import}
\usepackage{algorithm}
\usepackage{algorithmic}
\usepackage{amsfonts}
\usepackage{mathtools}
\usepackage{array}
\usepackage{longtable}
\usepackage{booktabs}	
\graphicspath{{fig/}}

\title{Episodic Reinforcement Learning with Expanded State-reward Space}

\author{
\hspace{-1mm}Dayang Liang,
Yaru Zhang and Yunlong Liu\footnotemark[1]\thanks{Corresponding Author}\\
Xiamen University \\
Xiamen, China\\
\texttt{ylliu @ xmu.edu.cn}
}

\iclrfinalcopy 

\begin{document}

\maketitle

\begin{abstract}
Empowered by deep neural networks, deep reinforcement learning (DRL) has demonstrated tremendous empirical successes in various domains, including games, health care, and autonomous driving. Despite these advancements, DRL is still identified as data-inefficient as effective policies demand vast numbers of environmental samples. Recently, episodic control~(EC)-based model-free DRL methods enable sample efficiency by recalling past experiences from episodic memory. However, existing EC-based methods suffer from the limitation of potential misalignment between the state and reward spaces for neglecting the utilization of (past) retrieval states with extensive information, which probably causes inaccurate value estimation and degraded policy performance. To tackle this issue, we introduce an efficient EC-based DRL framework with expanded state-reward space, where the expanded states used as the input and the expanded rewards used in the training both contain historical and current information. To be specific, we reuse the historical states retrieved by EC as part of the input states and integrate the retrieved MC-returns into the immediate reward in each interactive transition. As a result, our method is able to simultaneously achieve the full utilization of retrieval information and the better evaluation of state values by a Temporal Difference (TD) loss. Empirical results on challenging Box2d and Mujoco tasks demonstrate the superiority of our method over a recent sibling method and common baselines. Further, we also verify our method's effectiveness in alleviating $Q$-value overestimation by additional experiments of $Q$-value comparison.
\end{abstract}


Deep Reinforcement Learning~(DRL) algorithms have achieved human-level control with applications in various fields by combining feature extraction abilities of deep learning techniques and decision-making capabilities of reinforcement learning. Despite the great success, DRL requires extensive interactions to obtain sufficient reward signals and still suffers from sample inefficiency. For instance, even for the cutting-edge model-free Rainbow algorithm, to learn a human-level policy for video games, 18 million interaction steps are still required~\cite{ref_article1}. To address this issue, more dedicated methodologies, including exploration improvement~\cite{ref_article2,ref_article3}, environment modeling~\cite{ref_article4,ref_article5}, state abstraction~\cite{ref_article6,ref_article7,ref_article13}, and knowledge transfer~\cite{ref_article8,ref_article9}, are proposed.

Inspired by the memory-related hippocampus structure over brain decision-making~\cite{ref_article10}, to reuse the samples in the historic information, early research on model-free algorithms presented episodic control approaches to enable agents to perform appropriate decisions~\cite{ref_article11,ref_article12}. To be specific, the episodic control can achieve a reasonable estimation of the current value function by capturing the MC-returns of good state-action fragments in past memory. The idea is similar to the model-based DRL, like Dyna algorithm~\cite{ref_article17}, but only requires one or a few past samples to assist decision-making, while avoiding complex environment modeling and planning calculations. 

In the literature, many works have focused on how to effectively retrieve the historical information. Taking the system with discrete actions as an example, researchers have proposed efficient Gaussian random projection~\cite{ref_article15}, differentiable neural network~\cite{ref_article33} or K-means cluster~\cite{ref_article34} to compress the state-action pair into a low-dimensional form for the convenience of retrieval. However, once the past experiences have been retrieved,  the exploitations of the retrieval experiences, especially the information-rich historical states, are not well studied. In detail, these methods usually only exploit historical MC-returns to regularize the evaluation value of the current state, and then learn a comprehensive strategy that considers both current and historical factors, while the MC-returns' corresponding states that contain historical information are not taken into account as the input in the forward inference. Moreover, the propagating of past MC-returns~(weighted rewards) to the states without past information may lead to the misalignment of the state space and the reward space and the bias of the value estimation, the policy performance will be deteriorated.  Besides, if only the MC-returns in the retrieved information are utilized and as extensive computing resources are necessary for the retrieval process, current EC-based methods actually are sample inefficient.

To address the aforementioned issues, we propose an episodic control DRL method that expands the state-reward spaces, in which the spaces of input states and rewards are expanded by retrieval states and MC-returns, respectively. To achieve this, for expanding the states, we reuse the historical states retrieved by EC as part of the input states. For expanding the rewards, we directly integrate the retrieved MC-returns as part of the immediate rewards in a weighted manner during the TD loss calculation process~\cite{ref_article40}, while discarding the original auxiliary loss. Finally, both the states and rewards of the proposed method consist of a two-part space covering historical and current information. Compared to previous EC-based methods, the back-propagation approach under the new state-reward space enables the rewards to completely cover the input states containing past information, thus achieving better estimations of the state values and utilization of the retrieval samples.
 
We evaluate our approach on a set of challenging environments, i.e.,  Mujoco and Box2D tasks, and empirical experiments demonstrate the superiority of the algorithm compared to the strong baselines. Overall, our contributions are: 1) we present an episodic control-based DRL method with expanded state-reward spaces, where the spaces of training states and rewards are expanded by past states and MC-returns, respectively; 2) empirical experiments reveal that adopting an expanded state-reward space is beneficial to improve the policy performance and the utilization of retrieved states, which also mitigates the problem of value overestimation; 3) finally, we demonstrate the impact of different proportions between current and past information on decision-making through ablation experiments.

\section{Related Work}
\textbf{Model-based DRL}	In recent years, from the previous Dyna framework~\cite{ref_article16,ref_article17} to algorithms such as Stochastic Lower Bound Optimization~(SLBO)~\cite{ref_article18}, model-based reinforcement learning~(MBRL) methods have been developed rapidly with the advantages of high data utilization and strong portability due to the ability of predicting future information via learning an environment model. Although MBRL has many benefits and many classical methods have been proposed, for instance, the Model-Ensemble Trust-Region Policy Optimization~(ME-TRPO) framework~\cite{ref_article19}, a model-based algorithm that uses neural networks to model the dynamic environment and the Trust-Region Policy Optimization~(TRPO) to update the policy~\cite{ref_article20}, the performances of the related algorithms rely heavily on an accurate learned model, which is really hard to learn for complex systems in reality. To mitigate this problem, some solutions have been provided, such as Igl et al. ~\cite{ref_article21} trained the model and policy together, and Oh et al.~\cite{ref_article22} proposed a value prediction network based on end-to-end training manner, which uses the Monte Carlo algorithm to find the optimal action based the model learned by a neural network. Besides that, a Bayesian filters method is employed to model the environment~\cite{ref_article23}. 

\textbf{Model-free DRL}	In contrast with model-based algorithms, the model-free DRL methods generally learn the behavior policy directly by past data from a replay buffer, which avoids learning a complex model.   According to the policy update method, the model-free methods can be divided into the value-based and the policy-based gradient update. Take Deep Q-Network~(DQN)~\cite{ref_article24,ref_article25} as an example, DQN is an excellent value-based algorithm, which approximates the Q-value function through a neural network. For policy-based methods, such as Actor-Critic~(A2C) ~\cite{ref_article26}, Proximal Policy Optimization~(PPO) ~\cite{ref_article27} and TRPO, which have also achieved outstanding performance. Obviously, the direct use of buffer data is convenient and easy to implement. However, these model-free methods are generally limited to making decisions based on past historical information and lack the utilization of future information. Recent work, Imagination-Augmented Agents~(I2A) algorithm with imagination~\cite{ref_article28}, uses prediction trajectory generated by the imagination-augmented module, and then employs a model-free method to train the policy. This is similar to the model-based methods, where the predicted trajectory generated by an inaccurate model is often difficult to describe the actual action trajectory of the agent. 

\textbf{Episodic Control} To address the issues of model accuracy and sample inefficiency in both model-based and model-free methods, a model-free episodic control method pioneered by Blundell et al.~\cite{ref_article11} has been designed to purposefully recall past experiences while improving sample utilization of DRL. It introduces the concept of episodic memory into reinforcement learning and commonly uses a non-parametric $k$-NN search to recall past successful experiences quickly, the idea is similar to prioritized experience replay~\cite{ref_article30} but uses a separate mean square error loss to jointly optimize the Q-value function. In later improvements, on the one hand, Lin et al.~\cite{ref_article15} and Kuznetsov et al.~\cite{ref_article32} successfully applied episodic control to discrete and continuous control scenarios, respectively. On the other hand, Li et al.~\cite{ref_article33} and Liang et al.~\cite{ref_article34}  improved the inefficient episodic memory container via differentiable neural networks and $K$-means clusters, respectively. Although many episodic control-based approaches were proposed to improve the efficiency of DRL policy, they do not directly employ information such as real states in the trajectories. Consequently, due to the problem of the potential mismatch between state and reward spaces, some explored latent semantics, such as state transitions and topological similarities, cannot be explored and generalized well~\cite{ref_article41}. In this paper, we tackle this problem by realigning the state-reward space with historical retrieval information and improving the reward backpropagation method.

\section{Background}

The underlying system is modeled as a Markov Decision Process (MDP) defined by a tuple ${\mathcal M}=({\mathcal S}, {\mathcal A}, P, R, \rho_0,\gamma)$, where $\mathcal S$ denotes the low-dimension state space, $\mathcal A$ denotes the continuous action space, $ P(s_{t+1}|s_t, a_t){:\mathcal S} \times {\mathcal A} \to {\mathcal S} $ is the probability distribution function of transition from the state $s_t$ to the next state $s_{t+1}$ after taking an action $a_t$, $R(s_t, a_t):{\mathcal S} \times {\mathcal A} \to \mathbb{R}$ is the reward signal obtained by the system after taking an action $a_t$ in the state $s_t$, $\rho_0$ denotes the set of initial states, and $\gamma \in[0,1]$ is the discount factor of future rewards.

The aim of reinforcement learning is to learn a policy $\pi(a_t|s_t):{\mathcal S} \to {\mathcal A}$ for decision, which can be achieved by maximizing the expectation of the discounted cumulative reward, formulated as:
\begin{equation}
\label{eq1}
{\mathcal J}(\pi)=\mathbb{E}_{s_0\sim \rho_0,a_t\sim \pi(\cdot|s_t)\atop s_t\sim {\mathcal P}(\cdot|s_t,a_t)}\left[{\sum}^{\infty}_{t=0}{\gamma}^t R(s_t,a_t)\right].
\end{equation}

\subsection{Soft Actor Critic}\label{section:SAC}
Soft Actor Critic~(SAC) is a typical policy-based DRL algorithm for continuous control scenes~\cite{ref_article38}, which consists of an actor-network for learning the policy function $\pi_{\psi}(a_t|s_t)$ with parameters ${\psi}$ and a Critic network for learning the state-action value function $Q_{\phi}(s_t, a_t)$ with parameters ${\phi}$. Unlike value-based algorithms, SAC optimizes a stochastic policy to maximize the expectation of discounted cumulative reward. Additionally, compared with the basic Actor Critic~(AC) framework, the main improvement of SAC is that maximum entropy item ${\mathcal H}=log(\pi_{\psi}(\cdot|s_t))$ is added to decentralize the training policy, which enhances the exploration and robustness of the algorithm.

Concretely, SAC trains the Critic network via minimizing the Temporal-Difference~(TD) error~\cite{ref_article44}, i.e., the mean square error of approximate $Q$ and real $Q$ using current reward $r_t$. The training loss of parameters $\phi$ can be defined as follow:
\begin{equation}
\label{eq2}
{\mathcal L}^{Q}(\phi)=\mathbb{E}_{e_t\sim \mathcal B}\left[{\left(Q_{\phi}(s_t,a_t)-(r_t+\gamma (1-d){\mathcal T})\right)}^2\right].
\end{equation}
where $d$ represents the done signal, $e_t=(s_t,a_t,r_t,s_{t+1})$ is a transition data sampled from repaly buffer ${\mathcal B}=\{e_1,e_2,...,e_l\}$. The target $\mathcal T$ computes the expectation of the next actions sampling from current policy,  defined as:
\begin{equation}
\label{eq3}
{\mathcal T}=\mathbb{E}_{a_t\sim \pi}\left[\hat Q_{\hat{\phi}}(s_{t+1},a_t)-\alpha log(\pi_{\psi}(\cdot|s_{t+1})\right],
\end{equation}
where target network $\hat{\phi}$ comes from the Exponential Moving Average (EMA) of the Critic network parameter $\phi$, and the hyperparameter $\alpha$ is a positive entropy coefficient that determines the priority of entropy maximization over value function optimization.

Finally, we sample actions $a_t\sim\pi_\psi$ from the current policy, and train the Actor by maximizing the expected reward of the sampled actions:
\begin{equation}
\label{eq4}
{\mathcal L}^{\pi}(\psi)=\mathbb{E}_{a_t\sim \pi}\left[(Q^{\pi}(s_t,a_t)-\alpha log(\pi_{\psi}(\cdot|s_t))\right].
\end{equation}

\subsection{Gaussian Random Projection}
\label{section:GRP}
As mentioned, in our framework, we will use the current state-action pair to find similar state-action pairs from the Episodic Memory~(EM). However, since the state is in a high-dimensional form, the process of episodic retrieval needs to consume huge computing resources, which makes it difficult to achieve the expected function. To solve this problem, in this paper, Gaussian random projection is employed to build the index $h$ for every historical state-action pair to speed up the calculation. This method can guarantee that low-dimensional projection vectors retain most of the high-dimensional information.

Specifically, the theoretical basis of Gaussian random projection is based on Johnson-Lindenstrauss theorem~\cite{ref_article39,ref_article15}, which can be briefly summarized as follows. Given a data set with $N$ samples, such as $D = {\{X_i\}}^n_{i=1}$, where $X_i \in \mathbb{R}^d$ and $d$ is the dimension of $X_i$.  Given $0 \le \varepsilon \le 1$, then there exist a dimension-reduction mapping function: ${\mathcal F}:\mathbb{R}^d \to \mathbb{R}^m$, for $\forall~X_i,X_j \in D$:
	\begin{equation}
		\label{eq6}
		(1-\varepsilon)\parallel X_i-X_j \parallel \le  \parallel {\mathcal F}(X_i)-{\mathcal F}(X_j) \parallel \
		\le (1+\varepsilon)\parallel X_i+X_j \parallel
	\end{equation}
	
According to the above equation, even if the projection direction is selected by Gaussian random, as long as the dimension of the low-dimensional space satisfies certain conditions, the deformation caused by the projection operator to the original high-dimensional data is at most in the interval $[(1-\varepsilon),(1+\varepsilon)]$, i.e., the method can keep the structure~(the relationship between the data) of the data in the high-dimensional well.

\subsection{Episodic Control}
Episodic Control~(EC) is inspired by the mechanism of the brain's decision-making and motor control, i.e., the human brain leverages the striatum (focuses on current reflex) and hippocampus (focuses on past memory) to comprehensively complete a decision~\cite{ref_article10}. Here the EC simulates the influence of the hippocampus on decision-making of DRL. 

The key idea of EC is to utilize the current state-action pair to lock similar good policies in the past experience from Episodic Memory (EM), and these state-action pairs will be utilized to guide the learning of the current policy. Typically, EC computes the $L_2$ distance between the current transition and each historical transition in the EM, and then extracts the top $K$ historical transitions with the closest distances~\cite{ref_article11}. Next, we compute the mean Monte Carlo (MC)-returns $G_k$ along the subsequent trajectories of the $K$ transitions. $G_k$ is usually utilized to constrain the parameter $\phi$ learning of the current $Q$-value in the form of an auxiliary loss. Ultimately, the training loss  ${\mathcal L}^{EC}$ of the EC-based DRL algorithm can be summarized into the following common form:
\begin{equation}
\label{eq20}
\begin{matrix} {\mathcal L}^{EC}=\underbrace{\left(Q_{\phi}(s_t,a_t)-(r_t+\gamma (1-d){\mathcal T})\right)^2} + \\ current~term \end{matrix}  \begin{matrix}\underbrace{\lambda\left(Q_{\phi}(s_t,a_t)-G_k\right)^2} \\ past~term \end{matrix}
\end{equation}

In Equation (\ref{eq20}), the EC constrains the learning of the Q-value through current results $Q_{\phi}(s_t,a_t)$ and past results $G_k$, where the coefficient $\lambda$ is used to adjust the influence weight of the two terms. In our work, the proposed method adopts the past real states to train $Q$-value, instead of only using the MC-returns $G_k$, while discarding the auxiliary loss. More implementation details can be found in the section Method.

\section{Method}

\begin{figure*}[t]
\includegraphics[width=0.9\textwidth]{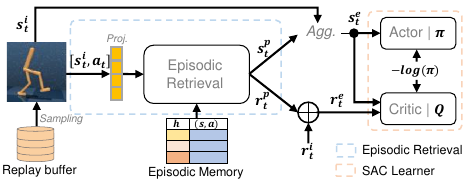}
\centering
\caption{Algorithm structure. The episodic control-based reinforcement learning approach with expanded state-reward space.} \label{fig:1}
\end{figure*}

In this section, we propose an EC-based DRL method with expanded state-reward space, which is extended to the Soft Actor-Critic (SAC) model-free DRL algorithm. As shown in Fig. \ref{fig:1}, our method primarily consists of two networks: the actor-network for learning policies and the critic-network for learning $Q$-value. Our goal is to reasonably utilize both the states and rewards information captured by episodic control to achieve better performances of the learned policy. 

\subsection{Overall Architecture}
As shown in Figure \ref{fig:1}, the architecture contains two crucial components, namely episodic retrieval (blue box, extended in Figure \ref{fig.2}) and standard SAC learner (pink box). Firstly, episodic retrieval is an upstream task aimed at retrieving similar past state-action pairs from the Episodic Memory~(EM) based on the low-dimensional random projection of the current $(s_t,a_t)$. In our improved retrieval module, it outputs these compact past states $s^p_t$  and the Monte Carlo returns $r^p_t$ of their subsequent trajectories, where $s^p_t \sim {\mathcal S}^p$, $r^p_t \sim {\mathcal R}^p$, and superscript `$p$' denotes `past'. Additionally, as for the downstream decision task, the $s^p_t$ and the current state $s^i_t$ (superscript `$i$' denotes `immediate') are spliced as the input state to train the actor-critic networks. Meanwhile, during TD learning, the integration of the  $r^p_t$ and the immediate environment reward $r^i_t$ is utilized to train the value network. Moreover, our method also requires two data containers: a replay buffer $\mathcal B$ for batch sampling and an EM $\mathcal M$  for episodic retrieval. Specifically,  $\mathcal M$ is a dictionary container, which stores the low-dimensional Gaussian projection $h_t$ (key) and the original transition $e_t$ (value) corresponding to the current $(s^i_t,a_t)$. $\mathcal B$ stores transition data with size $N$. Algorithm \ref{algorithm} shows the details of the overall framework.

\subsection{Episodic Retrieval}
\noindent\textbf{Gaussian Projection}~ In order to improve the efficiency of episodic retrieval, we search in low-dimensional projection space according to existing methods. The aforementioned Gaussian projection theorem shows that by multiplying the original matrix with the projection matrix, most of the features can still be preserved in low-dimensional vectors. Therefore, we use the projection method to project the state-action pair to any low-dimensional space, defined as $h_t={\mathcal G}([s^i_t;a_t]):{\mathcal S}^i\times {\mathcal A}\to  \mathbb{R}^n$, where $n$ represents the dimension of projection space.

\begin{figure*}[t]
\includegraphics[width=1\textwidth]{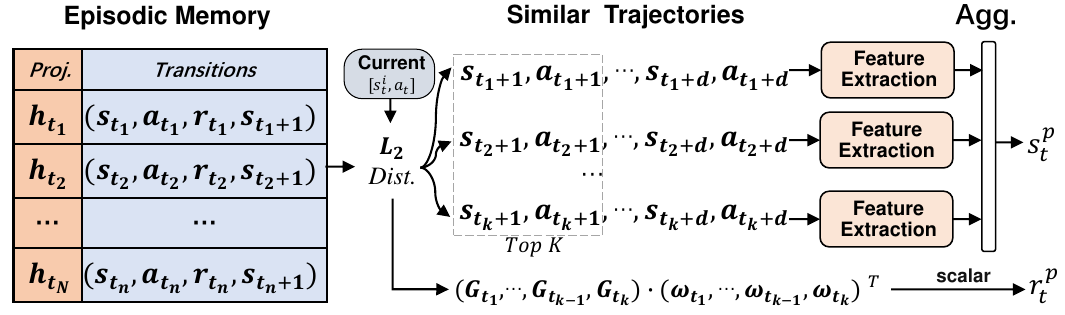}
\centering
\caption{Structure of episodic retrieval module} \label{fig.2} 
\end{figure*}

\noindent\textbf{Episodic Retrieval}~ As shown in Figure \ref{fig.2}, $\mathcal B$ with dictionary form holds all the original transitions and their projections before $t$ steps, formalized as,
\begin{equation}
		\label{eq7}
 {\mathcal B}=\{(h_{t_1}\colon e_{t_1}),(h_{t_2}\colon e_{t_2}),...,(h_{t_N}\colon e_{t_N}) \}.
\end{equation}

First of all, we introduce the retrieval process of the past state $s^p_t$. Given the projection vector of current $(s^i_t,a_t)$, we can calculate the distances from it to each projection vector in $\mathcal B$ by the $L_2$ distance, and then choose the $K$ transitions  $\{e_{t_1},e_{t_2},...,e_{t_K}\}$corresponding to the closest top $K$ distances. Then, the origin indexes of those transitions are used to extract the subsequent trajectory of each transition from buffer $\mathcal B$. Thereby, we express the set of the trajectories as $T=\{\tau_{t_1},\tau_{t_2},...,\tau_{t_K}\}$, where the length of each trajectory is set to $d$. To facilitate the implementation of the retrieval, the capacities of the $\mathcal M$  and the $\mathcal B$ need to be consistent. Ultimately, the $K$ trajectories (only pick up the state and action) are fed into the feature extraction networks and then aggregated into the past states $s^p_t$.

Secondly, we compute the MC-rewards for each trajectory in the set of $T$. According to the optional setting of trajectory length $d$, the MC-return can be divided into two cases:
\begin{align} \label{eq8}
\begin{split}
 G_{t}= \left \{
\begin{array}{ll}
    r_{t_1},                                      & if~~d=1,\\
    f^{\mathcal B}_d(s^i_t,a_t)     & if~~d>1,\\
\end{array}
\right.
\end{split}
\end{align}
where $f^{\mathcal B}_d(s^i_t,a_t)$ represents the approximate cumulative discounted return of a past trajectory retrieved by the current $(s^i_t,a_t)$. Actually, $d$ is set to 2 in this work, i.e., the MC-returns $G_t$ of a trajectory can be calculated by,
\begin{equation}
		\label{eq21}
G_t = f^{\mathcal B}_d(s^i_t,a_t)=\sum^{d}_{k=0}\gamma^k r_{t+k+1}.
\end{equation}

Then, we compute the MC-returns along each trajectory, expressed as the set of $\{G_{t_1}, G_{t_2},..., G_{t_K}\}$. Then, given the weight matrix $\{\omega_{t_1},\omega_{t_2},...,\omega_{t_K}\}$ that is obtained from the $L_2$ distances of the aforementioned $K$ transitions, we finally calculate the past reward $r^p_t$ by multiplying them, defined as:
\begin{equation}
\label{eq9}
r^p_t=(G_{t_1},...,G_{t_{k-1}},G_{t_K}) \cdot { (\omega_{t_1},...,\omega_{t_{k-1}},\omega_{t_K})}^\top.
\end{equation}

Finally, the retrieval module obtains two key historical information, i.e., the retrieved states $s^p_t$ and rewards $r^p_t$ through the above steps.

\subsection{Optimization  Implementation}
In the downstream tasks, the work still follows the inherent characteristics of the EC, yet we implement the same idea in a way that is highly data-efficient with stronger generalization ability. As a whole, our first direct improvement is to expand the state space of network training, where the learning of past states facilitates the generalization of learned policy to explored states. The other is that we try to incorporate the past reward into the immediate reward for the efficient learning of the value function.


Specifically, the input state information fed into the critic and actor networks is aggregated from the current state and past state, which can be expressed as:
\begin{equation}
\label{eq9}
 s^e_t=Agg(s^i_t,s^p_t).
\end{equation}

For the actual input state $s^e_t \sim {\mathcal S}^e$ (superscript `$e$' denotes `expanded') is defined on expanded state space ${\mathcal S}^e$, thus we accordingly introduce a comprehensive reward $r^e_t \sim {\mathcal R}^e$ defined on expanded reward space ${\mathcal R}^e$, calculated as:
\begin{equation}
\label{eq10}
 r^e_t= r^i_t + \eta ~  r^p_t
\end{equation}
where the weight coefficient $\eta$ between current and past rewards is adopted to fine-tune their influence on decision-making.

By properly introducing past information, we have obtained states $s^e$ and rewards $r^e$ that can be used for critic network training. Intuitively, the more reasonable state-reward space setting may lead to a good $Q$-value estimation, the formal analysis is in Section \ref{Matching Space}. At last, we train the critic network with an end-to-end loss, which is different from the work of Lin et.al~\cite{ref_article15} that regularizes $Q(s, a)$ with an auxiliary loss. The TD target is defined as,
\begin{equation}
\label{eq10}
{\mathcal Y} = r^e_t +\gamma (1-d) \mathbb{E}_{a_t\sim \pi}\left[\hat Q_{\hat{\phi}}(s^e_{t+1},a_t)-\alpha \mathcal H \right],
\end{equation}
and the critic-network is updated by
\begin{equation}
\label{eq11}
{\mathcal L}^{Q}(\phi)= \mathbb{E}_{(s^e_t,s^e_{t+1})\sim {\mathcal S}^e,r^e_t \sim {\mathcal R}^e, a_t \sim \mathcal B }\left[{\left(Q_{\phi}(s^e_t,a_t)-\mathcal Y\right)}^2\right],
\end{equation}
where $\mathcal H$ represents the maximum entropy learning item in the SAC algorithm, formulated in Section \ref{section:SAC}. The TD target is approximated by an expectation sampling method.

As mentioned, the policy $\pi(a_t|s_t)$ learned by the actor-network also depends on the expanded states $s^e_t$. According to the standard SAC algorithm~\cite{ref_article38}, the parameters $\psi$ of the actor can be learned by directly minimizing the KL divergence of the current policy distribution and $Q$-value, which is finally defined as:

\begin{equation}
\label{eq12}
{\mathcal L}^{\pi}(\psi)=\mathbb{E}_{s^e_t \sim {\mathcal S}^e, a_t\sim \pi}\left[(Q^{\pi}(s^e_t,a_t)-\alpha log(\pi_{\psi}(\cdot|s^e_t))\right].
\end{equation}
	
	    \begin{algorithm*}[t]
        \caption{Episodic Control-based DRL Method with Expanded State-reward Space}
        \label{algorithm}
        \begin{algorithmic}[1]
		\STATE Initialize replay buffer $\mathcal B$ and Episodic Memory  $\mathcal M$ with size $N$.
		\STATE Initialize critic networks $Q$, target $\hat{Q}$, and policy $\pi$  with parameters $\phi$, $\hat{\phi}$, and $\psi$, respectively.
		    \FOR {each iteration}                                     
		    	\FOR {each environment step}  
		    		\STATE $a_t \sim \pi_{\psi}(\cdot | s^e_t)$       \hfill {$\triangleright$ Sample an action from current policy.}
		    		\STATE $s^i_t,a_t,r^i_t,s^i_{t+1}=Step(a_t)$      \hfill {$\triangleright$ Execute an action.}
 					\STATE ${\mathcal B} \gets {\mathcal B}\ \cup\ <s^i_t,a_t,r^i_t,s_{t+1}> $         \hfill{$\triangleright$ Collect a transition.}
                	\STATE $h_t={\mathcal G}([s^i_t;a_t])$                                            \hfill{$\triangleright$ Compute  Gaussian projection vector.}
					\STATE ${\mathcal M} \gets {\mathcal M}\ \cup\ <h_t,s^i_t,a_t,r^i_t,s^i_{t+1}>$     \hfill{$\triangleright$ Collect an episodic memory.}
                \ENDFOR                                                        
                    \FOR{each gradient step}                               
                    \STATE  $(h_t,s^i_t,a_t,r^i_t,s^i_{t+1}) \sim U({\mathcal B})$   \hfill{$\triangleright$ Randomly sample a batch of samples.}
                    \STATE Retrieve the past information $s^p_t$ and reward $r^p_t$ of ${\mathcal G}([s^i_t;a_t])$ by EC.
                    \STATE $ s^e_t=Concat(s^i_t,s^p_t)$;  $ r^e_t= r^i_t + \eta ~  r^p_t$  \hfill{$\triangleright$ Compute the expanded state and reward.}
                    \STATE $\hat{\nabla}{\mathcal L}^Q(\phi)=\nabla_\phi Q_\phi(a_t,s^e_t)\left(Q_\phi(a_t,s^e_t)-(r^e_t+\gamma Q_{\hat{\phi}}(a,s^e_t)-\alpha \mathcal H)\right)$   \hfill{$\triangleright$ Update the gradients of critic network.}
		    \STATE $\hat{\nabla}{\mathcal L}^\pi(\psi)= \nabla_\psi log(\pi_\psi(\cdot|s^e_t))$  \hfill{$\triangleright$ Update the gradients of actor network.}
		    \STATE $\hat{\phi} \gets \tau\phi+(1-\tau)\hat{\phi}$                         \hfill{$\triangleright$ Update the target critic network.}
               	    \ENDFOR	                                                        
               	   \ENDFOR         
        \end{algorithmic}
    \end{algorithm*}
    
\subsection{Space Alignment}
\begin{figure}[h]
\includegraphics[width=0.7\textwidth]{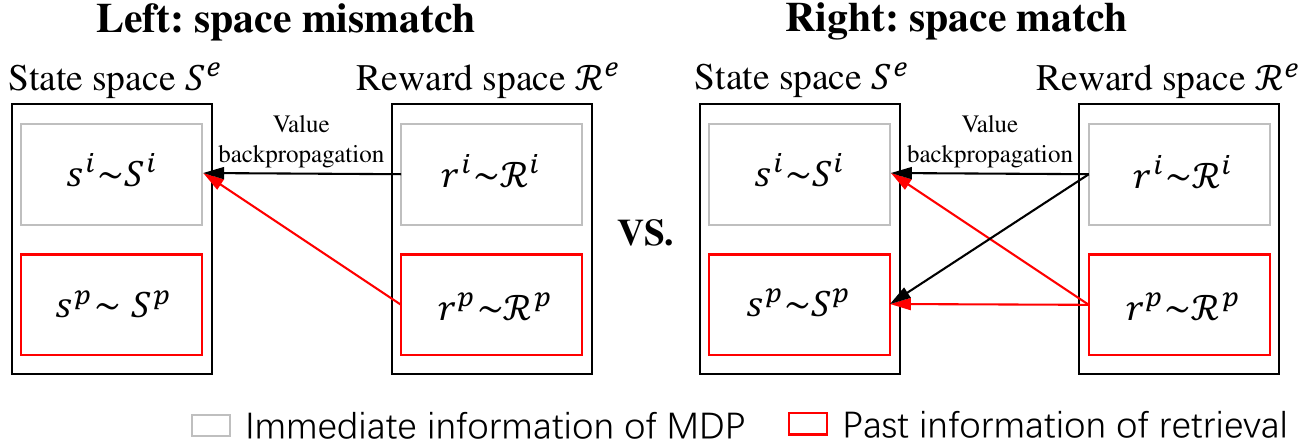}
\centering
\caption{Two match relationships between state and reward space during value back-propagation.} \label{fig.3} 
\end{figure}

In episodic control-based DRL, we can formalize the expanded state (reward) model space as the concatenation of the immediate and past state (reward) spaces, i.e., ${\mathcal S}^e={\mathcal S}^i+{\mathcal S}^p$ and ${\mathcal R}^e={\mathcal R}^i+{\mathcal R}^p$, as shown in Figure \ref{fig.3}. In training, $r^e \sim {\mathcal R}^e$ is leveraged to learn $V(s)$ under state $s^e\sim {\mathcal S}^e$, where the absence of any subspace of $\mathcal S$ and $\mathcal R$ may lead to a mismatch between the reward and the state transition model space \cite{ref_article45}, which in turn leads to the problems of distorted estimation and policy generalization on back-propagation period of $r^e \to s^e$. In Figure \ref{fig.3} (left), past rewards are back-propagated to a current state that does not produce that reward, which in fact describes a potential mismatched learning pattern in episodic control methods. In our work, the proposed method has the property of alleviating this problem. Specifically, we abstract past information $s^p$ as part of the actual input state  $s^e$, while incorporating past MC-rewards (i.e., $r^p$) into the internal computation of the TD target, which guides value update with a clear alignment mode.

\label{Matching Space}

\section{Experiments}
We evaluate the proposed algorithm on a series of challenging Box2d and Mujoco physics tasks. The goals of the experimental evaluation are as follows: 1) Demonstrate the performance improvement of the proposed algorithm compared to the strong sibling EMAC baseline and common baselines; 2) Empirically reveal that the proposed algorithm with expanded state-reward space has the ability to further alleviate the problem of $Q$-value overestimation; 3) Through ablation experiments show how to achieve the best trade-off between the past and the current information for learning the best policy, as well as shows the best coefficient of different tasks. 

All algorithms utilize a typical Adam optimizer~\cite{ref_article43}with the same batch size of 256, and the learning rates of 3e-4 (TD3) or 1e-3 (others). The models' networks consist of two hidden layers, size 256, for the actor and the critic, and a rectified linear unit (ReLU) as a nonlinearity. The projection dimension is set to 10, the typical coefficient $\eta$ is 0.5~(0.25 in \textit{Hopper-v3}, 1.0 in \textit{LunarLander-v2}), the discount reward is 0.99, the initial temperature coefficient is 0.1 and so on. Detailed parameters can be accessed in Table~\ref{tab:2}.

\subsection{Environments}

\begin{figure*}[b] 
\includegraphics[width=1\textwidth]{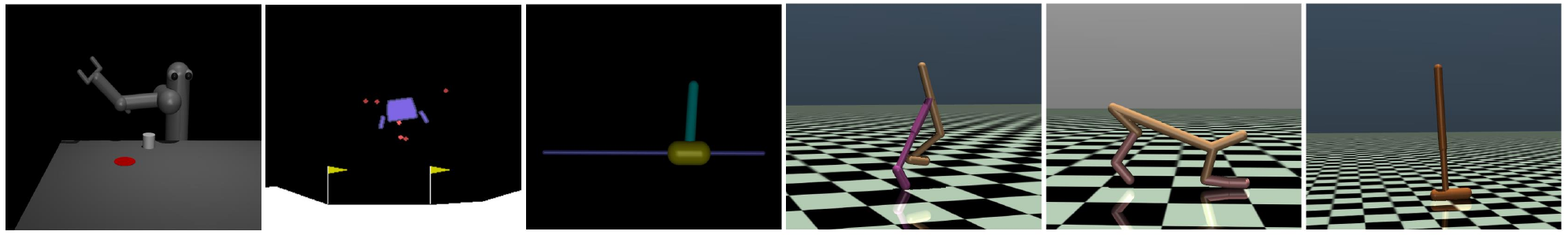}
\centering
\caption{ Illustrations of the experimental environments. From left to right: \textit{Pusher-v2, LunarLanderContinuous-v2, InvertedPendulum-v2, Walker2d-v3, HalfCheetah-v3, and Hopper-v3}.} \label{fig.4} 
\end{figure*}

The environments in Figure \ref{fig.4} are briefly described below. \textit{Pusher-v2}: the task is to move the stacking to the red position by controlling the 4-DOF robotic arm; \textit{LunarLander-v2}: the task is to control the three jets of the satellite to successfully land on the designated position; \textit{InvertedPendulum-v2}: its task is to balance the connecting rod by controlling the slider; \textit{HalfCheetah-v3}: the task is to learn to run by controlling the multi-degree-of-freedom cheetah; \textit{Walker2d-v3} and \textit{Hopper-v3}: the tasks are to control the foot-shaped agent to learn to walk and stand, respectively.

In this work, the training states are the low-dimensional vector of aforementioned tasks, such as joint angle, acceleration, and position, as well as the controlled actions are generally continuous angle, speed, or direction.

\subsection{Main Result}
\begin{figure*}[t]
\includegraphics[width=1\textwidth]{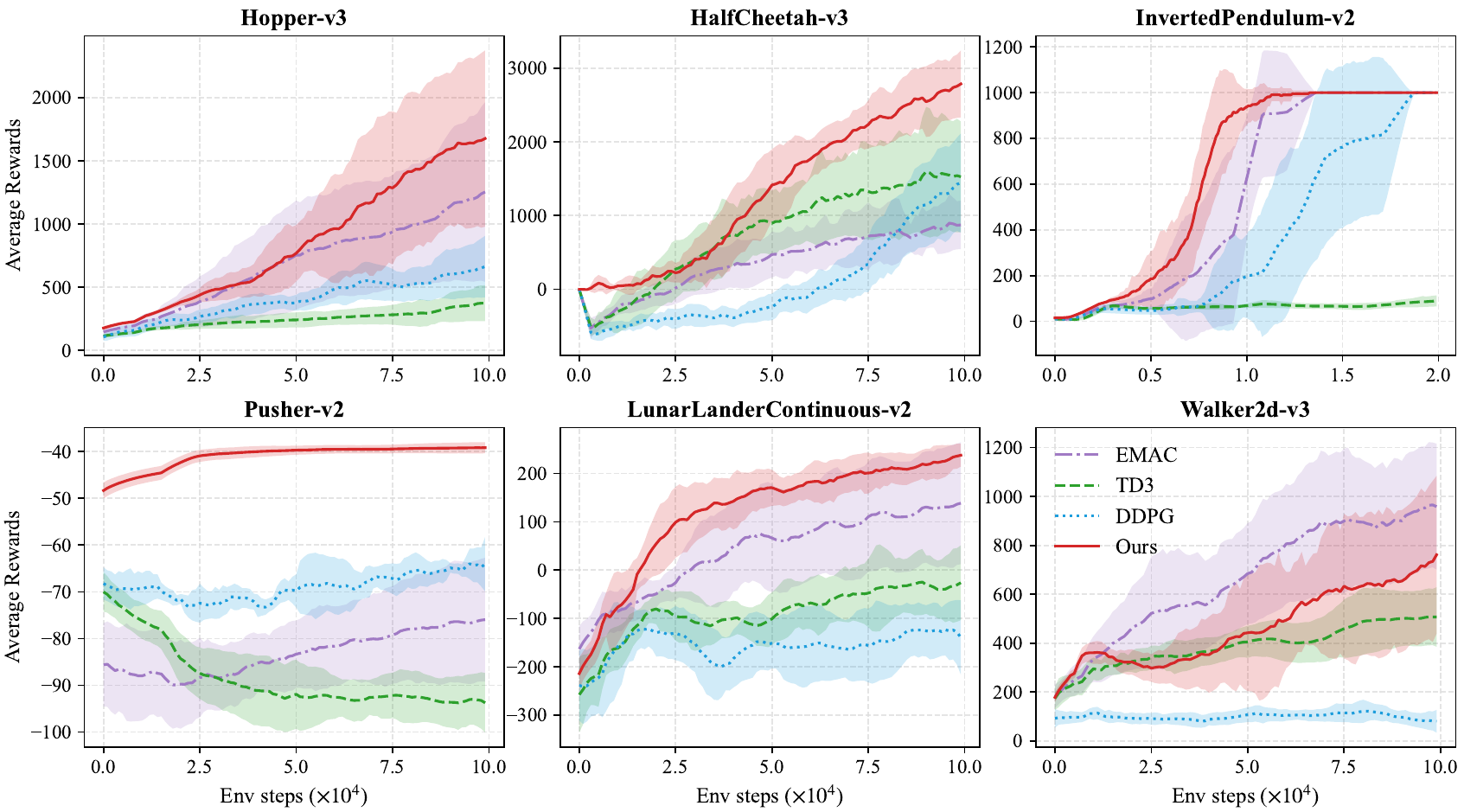}
\centering
\caption{Performance comparison for 100K environment steps (20k steps in \textit{InvertedPendulum-v2}) on Mujoco and Box2d tasks. For every curve, the mean episode rewards are computed every 1000 environment steps (100 steps in \textit{InvertedPendulum-v2}), averaging over 10 episodes. Each curve is averaged over 10 seeds and is smoothed for visual clarity.} \label{fig.5} 
\end{figure*}

As shown in Figure \ref{fig.5}, we compare our method with a recent sibling algorithm, i.e., EMAC~\cite{ref_article32} and various variants of the Soft Actor-Critic, including the powerful TD3~\cite{ref_article42} in the continuous control field, and the common DDPG~\cite{ref_article31} baseline. Among them, the sibling EMAC also aims to exploit past good experiences and share a similar idea with episodic control, which is the key comparison baseline of our method. 

\begin{table*}[t] \scriptsize
    \centering
    \caption{Comparison results of the best mean episode rewards on the 6 tasks. (mean $\&$ standard deviation for 10 seeds)}
    \label{tab:1}
    \begin{tabular}{lcccccc}
        \toprule
        \textit{Methods}      &\textit{Hopper-v3}  & \textit{HalfCheetah-v3} & \textit{InvertedPendulum-v2} & \textit{Pusher-v2} & \textit{LunarLanderC-v2} & \textit{Walker2d-v3}\\
        \midrule
        EMAC         & 1254$\pm$710 & 898$\pm$393 & 1000$\pm$0 & -76$\pm$13 & 139$\pm$126 &  \textbf{966$\pm$255} \\
        TD3           & 374$\pm$141 & 1597$\pm$879 & 90$\pm$24 & -70$\pm$4 & -25$\pm$68 &507$\pm$117 \\
        DDPG        & 664$\pm$243 & 1442$\pm$672 & 1000$\pm$0 & -64$\pm$3 & -122$\pm$36 & 122$\pm$41 \\
        Ours           & \textbf{1677$\pm$697} & \textbf{2788$\pm$455} & \textbf{1000$\pm$0} & \textbf{-39$\pm$1} & \textbf{238$\pm$24} & 761$\pm$322 \\
        \bottomrule
    \end{tabular}

\end{table*}

The environments for evaluating algorithm performance cover Mujoco environment~\cite{ref_article46} (\textit{Hopper-v3, HalfCheetah-v3, Walker2d-v3} tasks) and Gym-Box2d environment (\textit{InvertedPendulum-v2, Pusher-v2, LunarLander-v2} tasks). As shown in Table \ref{tab:1}, our method achieves the best average rewards and faster convergence in most tasks, which means our agent is able to get a better policy performance over long-term interactions. Specifically, even in the simple task of \textit{InvertedPendulum-v2} where the convergence takes only 10k steps, our method can still improve the learning efficiency again with such a narrow improvement space. In addition, in the \textit{HalfCheetah-v3} task, our method significantly improves by nearly 75\% over the best baseline. However, note that our algorithm performs poorly in the hard\textit{Walker2d-v3} task. We suspect that this may be due to the sparse reward property of high-difficulty tasks, which forces our method to still use an almost constant reward signal while expanding the state space, thereby increasing the burden of state space exploration and value learning. Overall, the presented method outperforms the sibling EC-based EMAC algorithm, while also significantly outperforming other baselines.

\subsection{Q-value Overestimation}

\begin{figure*}[th]
\includegraphics[width=1\textwidth]{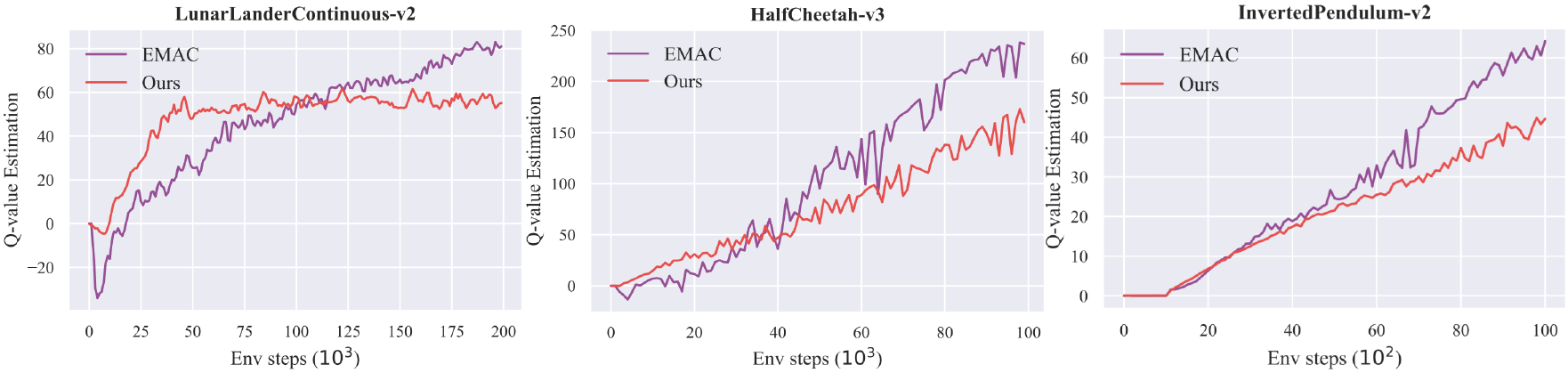}
\centering
\caption{Q-value estimation comparison between the proposed method and EMAC under the same setting.  Each curve is averaged over 5 seeds and not smoothed. } \label{fig.7} 
\end{figure*}

$Q$-value overestimation is a main challenge in the literature and for the inherent overestimation of the Actor-Critic~ (AC) framework, the problem can be traced back to the Q-learning algorithm~\cite{ref_article25}, i.e., the TD target of the Q-learning algorithm is derived from the optimal Bellman equation, which leads to an overestimation bias of the $Q$-function during value back-propagation. Although our approach is also based on the AC framework, we show that we can alleviate the problem well both theoretically and experimentally. 

From the theoretical perspective, since the overestimation problem has been well addressed in EC-based methods such as EMAC, which has the ability to alleviate the overestimation problem caused by the AC framework~\cite{ref_article15,ref_article32}, our EC-based approach theoretically can have the same performances in terms of the Q-value estimation. Furthermore, as shown in  Figure~\ref{fig.7}, with the alignment of the state-reward spaces, we show experimentally the problem can be further mitigated and the $Q$-value estimation of our method is significantly lower than that of the EMAC algorithm.

\subsection{Ablation Study}

\begin{figure*}[b]
\includegraphics[width=1\textwidth]{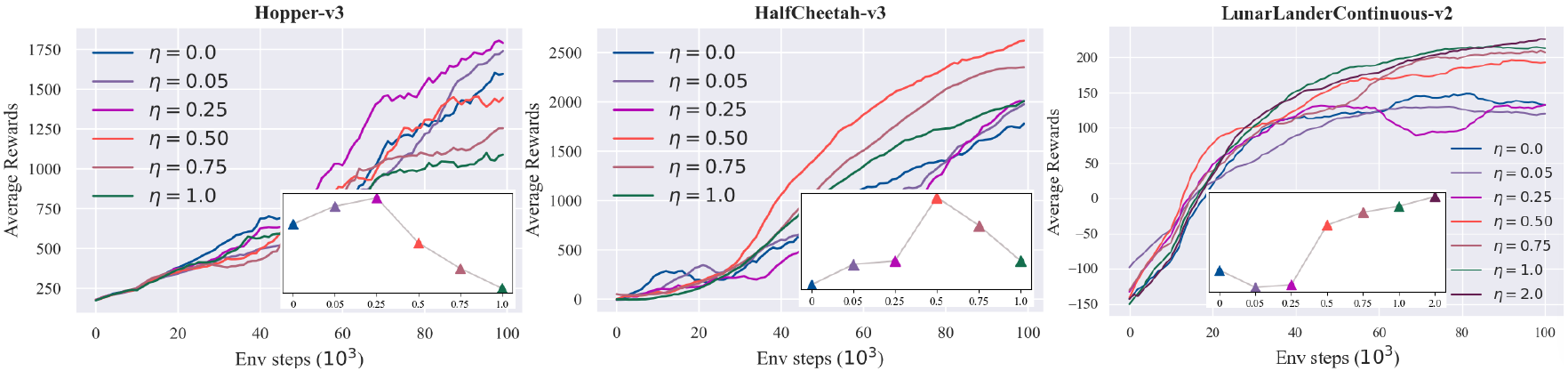}
\centering
\caption{Ablation experiments on several tasks. We test the performances of the proposed method by tuning a series of trade-off coefficients $\eta$. Each curve is averaged over 5 seeds. The embedded line graph represents the best average rewards corresponding to the ablation curves of each task. Note that coefficient $\eta=2.0$ is additionally supplemented in LunarLanderContinuous-v2.} \label{fig.6} 
\end{figure*}

Empirically, human beings can always make a comprehensive decision on the current event through experience results and current state. Specifically, they tend to rely on feedback from past experiences when dealing with memory-type events. On the contrary, in reflection-type events, they are used to making decisions based on immediate feedback. In our method, we adopt the weight coefficient $\eta$ between current and past results to quantify their influence on the decision.

To explore the optimal decision-making scheme, we conduct ablation experiments to reveal the performance when fine-tuning the weight between the current and past rewards. Therefore, we set a set of typical coefficients $\eta=\{0.0, 0.05, $ $0.25, 0.5, 0.75, 1.0\}$ for the ablation, where the two extreme values mean that only immediate or half-and-half mixture reward is leveraged to guide policy learning.

As shown in Figure \ref{fig.6}, we perform ablation experiments on the \textit{Hopper-v3}, \textit{HalfCheetah-v3}, and \textit{LunarLanderC-v2} tasks. By the comparison results, we can summarize the following empirical conclusions: 1) Fine-tuning $\eta$ will allow agents to learn policies with different performances. Nevertheless, making decisions that rely exorbitantly on immediate($\eta=0.0$) or past feedback, i.e., reward ($\eta=1.0$) is usually not the best solution, which is broadly consistent with practical experience. 2) Making a good trade-off between immediate and historical rewards is beneficial for the agent to learn a good policy; 3) The optimal coefficients for different tasks are not fixed, yet generally in the interval of $\eta \in(0,1.0)$. For example, as shown in the embedded line graphs, in \textit{Hopper-v3}, $\eta=0.25$ achieves the best performance, while in the \textit{LunarLanderContinuous-v2} task $\eta=2.0$.
\begin{table}[ht]
    \centering
    \caption{Hyperparameters used for experiments}\label{tab:2}
    \begin{tabular}{ll}
        \toprule
        \textit{Hyperparameter}        & \textit{Value}  \\  
        \midrule
        Seeds           & 0-9           \\
        Replay buffer size    & 100000           \\
        Episodic memory size    & 100000           \\
        Coefficient $\eta$       & 0.25,~~Hopper-v3         \\
                            & 1.0,~~LunarLanderC-v2        \\
                             & 0.5,~~otherwise        \\
        Environment steps       & 20000,~~InvertedPendulum-v2         \\
                            & 100000,~~otherwise        \\
        Evaluation interval       & 100,~~InvertedPendulum-v2         \\
                            & 1000,~~otherwise        \\
         Batch Size         & 256           \\
          Discount factor & 0.99 \\	 
         Hidden layer size       &256           \\        
          K-nearest size  & 2           \\
          Memory dimension & 10 \\
         Optimizer         & Adam           \\
          Learning rate  & 0.001           \\
        \bottomrule
    \end{tabular}
\end{table}
\section{Conclusion}
We introduce an episodic control approach with expanded state-reward space under model-free DRL algorithms that is able to improve policy performance. In our method, the valuable retrieval states are reconsidered as part of the training states, while the retrieved MC-returns are directly integrated as part of the immediate rewards in a weighted manner during the TD loss calculation. Thus, both the state and reward of our method consist of a two-part space covering historical and current information. Ultimately, our method can achieve the full utilization of retrieval information, while achieving better evaluation of state value through a space-aligned training manner.

Experiments are conducted on challenging Box2d and Mujoco continuous control tasks, and the main results show that the performance of our method has obtained improvement compared with authoritative baselines. Besides, comparison results of the $Q$-value show that the proposed method is able to effectively alleviate the problem of $Q$-value overestimation. Finally, ablation experiments also show that a reasonable trade-off between historical and current results is conducive to learning an optimal policy. 

\section*{Acknowledgments}
This work was supported by the National Natural Science Foundation of China (No. 61772438 and No. 61375077).

%


\bibliographystyle{iclr2016_conference}
\bibliography{sample}

\end{document}